\documentclass{article}

\usepackage{style}
\usepackage[utf8]{inputenc}
\usepackage{cite}
\usepackage{amsmath,amssymb,amsfonts}
\usepackage{algorithmic}
\usepackage{graphicx}
\usepackage{multirow}
\usepackage{textcomp}
\usepackage{relsize}
\usepackage{mathtools}
\usepackage{enumitem}
\usepackage{abstract}
\usepackage{array}
\usepackage{multirow,multicol}

\title{\textbf{Entity Profiling in Knowledge Graphs}}
\author{
\textbf{Xiang Zhang$^{1,4}$}
\thanks{Corresponding author: x.zhang@seu.edu.cn}
\thanks{This work was supported by the National Natural Science Foundation of China under grant U1736204, and the National Key Research and Development Program of China under grant 2017YFB1002801, 2018YFC0830201. This paper is partially funded by the Judicial Big Data Research Center, School of Law at Southeast University.}
\thanks{\copyright 2020 IEEE. Personal use of this material is permitted. Permission from IEEE must be obtained for all other uses, in any current or future media, including reprinting/republishing this material for advertising or promotional purposes, creating new collective works, for resale or redistribution to servers or lists, or reuse of any copyrighted component of this work in other works.}\and 
\textbf{Qingqing Yang$^2$}\and 
\textbf{Jinru Ding$^3$}\and 
\textbf{Ziyue Wang$^4$}\and\\
\affiliations
$^1$School of Computer Science and Engineering, Southeast University, Nanjing, China,\\
$^2$Southeast University - Monash University Joint Graduate School, Suzhou, China,\\
$^3$School of Software Engineering, Southeast University, Suzhou, China,\\
$^4$School of Cyber Science and Engineering, Southeast University, Nanjing, China\\
\emails
\{x.zhang, yangqing, dingjinru, wangzy1130\}@seu.edu.cn}

\date{}

\begin{document}
\maketitle
\begin{abstract}
\noindent Knowledge Graphs (KGs) are graph-structured knowledge bases storing factual information about real-world entities. Understanding the uniqueness of each entity is crucial to the analyzing, sharing, and reusing of KGs. Traditional profiling technologies encompass a vast array of methods to find distinctive features in various applications, which can help to differentiate entities in the process of human understanding of KGs. In this work, we present a novel profiling approach to identify distinctive entity features. The distinctiveness of features is carefully measured by a HAS model, which is a scalable representation learning model to produce a multi-pattern entity embedding. We fully evaluate the quality of entity profiles generated from real KGs. The results show that our approach facilitates human understanding of entities in KGs. 
\end{abstract}

\section{Introduction}
\label{sec:introduction}
Recent years witnessed a rapid growth in knowledge graph (KGs) constructions. Many KGs have been created and applied to real-world applications. A KG stores factual information in the form of relationships between entities and attributes of entities. In tasks like entity searching \cite{1} or data integration \cite{2}, users need to investigate entities quickly and frequently. Comprehension of an entity involves two types of user understanding: one is to identify an entity to its corresponding real-world object; the other is to compare an entity among others to understand its uniqueness. The volume and the structural complexity of KG significantly decrease the efficiency in identifying or comparing entities. To mitigate the problem of entity identification in entity comprehension, a research area of entity summarization has emerged in recent years, as stated in \cite{3}. The approach of entity summarization tries to shorten a lengthy entity description by extracting a concise summary, and to preserve informative statements in the summary.

While extracting summaries can help users quickly identify entities in interest, the problem to distinguish a given entity among others is still unsolved. It is not easy for users to discover the uniqueness of an entity by a self-describing summary. We often face this kind of situations in real life. Is someone active in his social network? Is a movie rated 2 stars worth watching? These questions have to be answered with the comparative information of other social actors or movies. User's view on an entity changes when comparing the entity to other similar entities. Comparing entities gives users a deep understanding of distinctive features of an entity. 

The distinctiveness of entity features cannot be presented in extractive summaries. It is because these summaries only encompass ``local" information of entity itself, but lack of ``global" information of how the entity shows uniqueness comparing to others. In this paper, we present an abstractive approach to profile entities in KG. Structure labels representing distinctive entity features will be abstracted from a KG by graph analysis. An entity profile is a short list of labels that the entity matches, which shows its prominent features.

\begin{figure}[h!]
\centering
\includegraphics[scale=0.45]{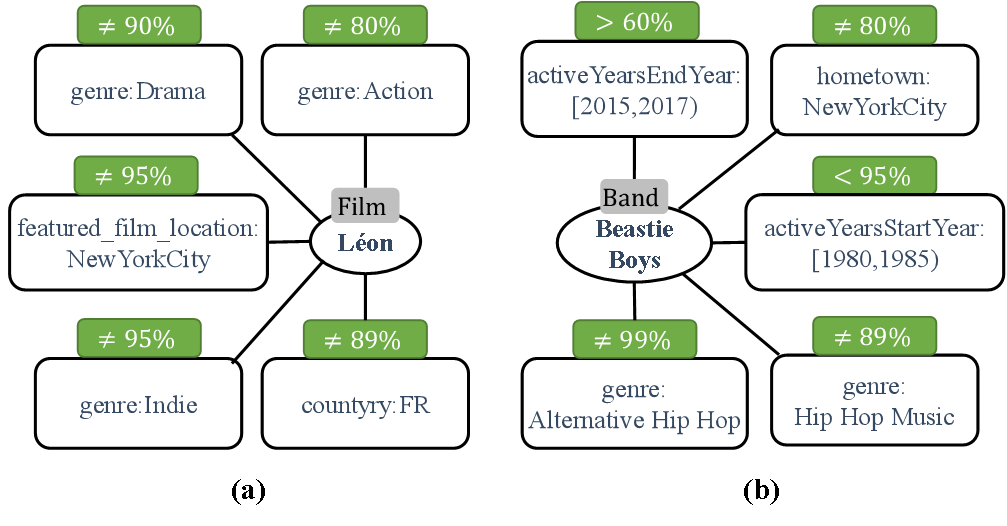}
\caption{Two entity profiles: (a) the film $L\acute{e}on$ in LinkedMDB; (b) the band $Beastie Boys$ in DBpedia.}
\label{fig1}
\end{figure}

We present a visualized example of two entity profiles generated by our approach in Fig.\ref{fig1}. One of the entities we select is $L\acute{e}on$, which is a film entity defined in the movie KG LinkedMDB\footnote{LinkedMDB: http://www.linkedmdb.org/}; the other is $Beastie Boys$, which is a Band entity defined in DBpedia \cite{4}. Each entity is profiled with five labels.  Each label is extracted from the KG, and is attached with an distinctiveness indicator (in green), where ``$\neq$ 80\%" means that this entity is different with 80\% of the other films or bands in this certain feature; ``$>$ 60\%" or ``$<$ 95\%" indicates that this entity has a larger or smaller value comparing to a proportion of other films or bands in this certain feature.
\vspace{0.5mm}

The main contributions of this work are in particular: (1) We propose the problem of entity profiling in KG, and we provide the first solution to this problem (to the best of our knowledge); (2) We propose a HAS model, which is a scalable and multi-pattern representation learning model. In our scenario, HAS model is used to efficiently find most distinctive labels in a KG for entity profiling. (3) We carry out an extensive empirical study of the proposed approach. Both intrinsic and extrinsic evaluations show that our approach facilitates human-understanding of the uniqueness of entities. 

The remainder of this paper is organized as follows. The problem is stated in Section \ref{sec:problem}, as well as an overview of our approach to solving the problem. We discuss how we generate candidate labels in Section \ref{sec:generate}. The measurement of distinctiveness for each candidate label, together with the profile generation, is detailed in Section \ref{sec:measure}, where we propose the HAS model. Intrinsic and extrinsic evaluations are given in Section \ref{sec:evaluation}, as well as the discussion. Related works are discussed in Section \ref{sec:related works}. We conclude our approach in Section \ref{sec:conclusions}.

\section{Problem Statement}
\label{sec:problem}
In this section, we give a definition of the problem we investigate, and then explain the flow of our approach. 

\vspace{0.1cm}
\textbf{Definition 1 (Knowledge Graph).} A Knowledge Graph $\mathcal{G}=\left \langle \mathcal{V,U,}\boldsymbol{\tau,\mu}\right \rangle$, where $\mathcal{V}=\mathcal{E} \cup \mathcal{L}$ is the set of nodes in $\mathcal{G}$. $\mathcal{E}$ is the set of entities; $\mathcal{L}$ is the set of literals;  $\mathcal{U}$ is the set of edges, and each edge connects an entity to another entity in $\mathcal{E}$ or a literal in $\mathcal{L}$;  $\boldsymbol{\tau}$ is a typing function on entities mapping each entity to one or more pre-defined types; $\boldsymbol{\mu}$ is an edge-labelling function, mapping each edge to a property.

\vspace{0.1cm}
\textbf{Definition 2 (Label and Label Set).} Given a type $t$ in $\mathcal{G}$, a label set $\mathbb{L}_t$ is a finite set of labels describing features of type $t$. Each label in $\mathbb{L}_t$ is a triple: $l=\left \langle t, l_{property}, l_{value} \right \rangle$. $l_{property}$ refers to a property, which is a distinctive attribute or a relation an entity may possess, and $l_{value}$ is the value of the property, the structure of which will be discussed in the next section.

\vspace{0.1cm}
\textbf{Definition 3 (Entity Profiling).} Given a knowledge graph $\mathcal{G}$, entity profiling is a two-step process: (1) For each type $t$ in $\mathcal{G}$, a label set $\mathbb{L}_t$ will be automatically abstracted; (2) For each entity $e$ of type $t$, a profile of $e$ is generated as: $profile(e)=\left\langle l_1, l_2,\ldots ,l_m \right\rangle$, which is an ordered set of labels, and $l_i \in \mathbb{L}_t$.

\vspace{0.1cm}
The core idea in entity profiling is to construct a label set for each type of entities. In many scenarios of profiling, such as user profiling \cite{5} or data profiling \cite{6}, distinctive labels are usually pre-determined and carefully selected by human experts driven by extrinsic business objectives or requirements of analytical practice. In entity profiling, without prior knowledge of the entities to be investigated, the label set of profiling has to be determined automatically. 

\begin{table}[h]
\caption{Categorization of Labels and Corresponding Examples}
\vspace{2mm}
\setlength{\tabcolsep}{2pt}
\begin{tabular}{|p{35pt}|p{208pt}|}
\hline
\textbf{Labels} & \textbf{Examples} \\
\hline
\textit{AIL} & $\left\langle Film,rating,[8.0,9.0]\right\rangle$\\
\textit{AVL} & $\left\langle Person,gender,``female"\right\rangle$ \\
\textit{RAL} & $\left\langle Director,directorOf,\left\langle Film,rating,[8.0,9.0]\right\rangle\right\rangle$\\
\textit{REL} & $\left\langle Product,producer,Apple\right\rangle$ \\
\hline
\end{tabular}
\label{tab1}
\end{table}

Features of entities are heterogeneous in structure. Some features are attributive, describing entities with attributive values, such as the age group or educational level of a person, or the rating of a film. Other features are relational, showing distinctive connections between entities, such as a famous film director who directed a number of highly-rated films. To help users fully understand the uniqueness of an entity with a comprehensive profile, we design a categorization on labels based on their structures, as shown in Table \ref{tab1}. \textit{AIL} and \textit{AVL} are two subsets of attributive labels. \textit{AIL} stands for attributive-interval labels, which are features showing that a certain attribute of entities falls in a prominent value interval. For example, the label $\left\langle Film, rating, [8.0, 9.0] \right\rangle$ describes films that are highly-rated and usually worth-watching. \textit{AVL} stands for attributive-value labels, where the value of the label is not a interval but specified. $\left\langle Person, gender, ``female" \right\rangle$ is an \textit{AVL}. Besides, \textit{RAL} and \textit{REL} are the two subsets of relational labels. \textit{RAL} stands for relational-attributive labels, which shows that some entities have connection to others with certain attributes. For example, $\left\langle Director, directorOf, \left\langle Film, rating, [8.0, 9.0] \right\rangle \right\rangle$ describes the group of Directors who have directed high-rated Films. \textit{REL} stands for features that connecting entities to a certain entity. For example, iPhone, iPad and other Apple products are all entities with the label: $\left\langle Product, producer, Apple \right\rangle$.

\vspace{1mm}
\begin{figure}[h!]
\centering
\includegraphics[scale=0.5]{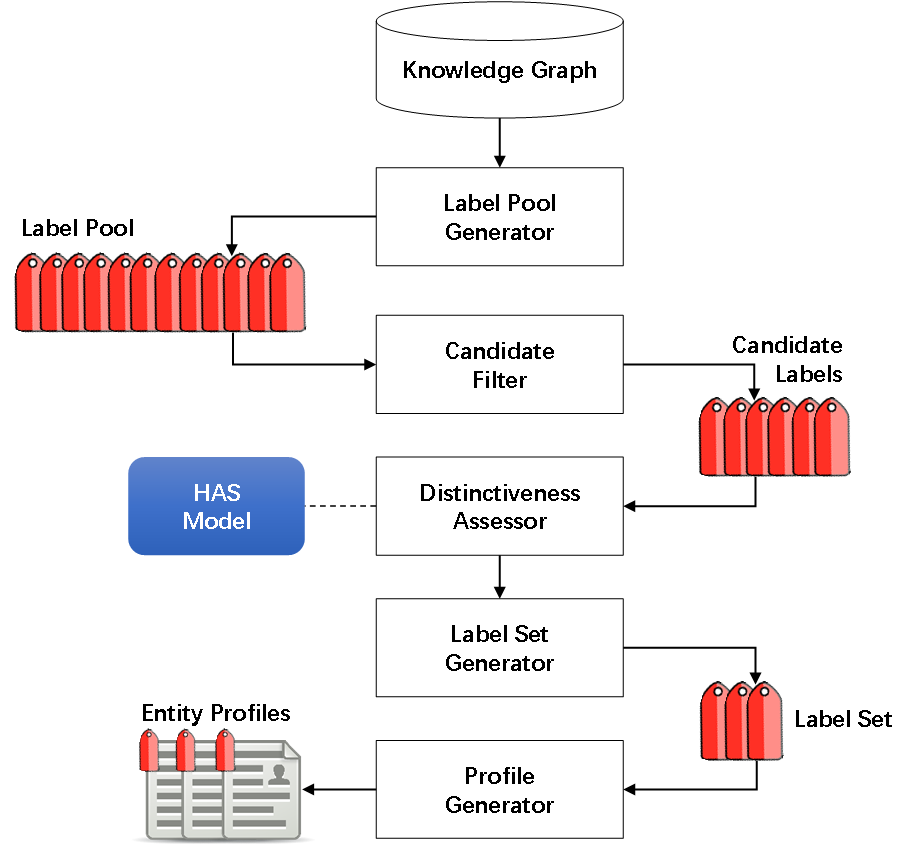}
\caption{The work flow of entity profiling.}
\label{fig2}
\end{figure}

The process of our approach is illustrated in Fig.\ref{fig2}. First, given a KG as input, all potential labels are automatically enumerated into a Label Pool. The enumeration may result in a huge number of candidate labels. These candidates will go through a preliminary heuristic filtering process by Candidate Filter. Apparently indistinctive features will be discarded. After that, each candidate label will be thoroughly examined by Distinctiveness Assessor. A HAS model is proposed to measure the difference between the positives and negatives for any given label. Only distinctive labels are enrolled in the Label Set, and we use a re-ranking to reduce redundancy in the label space. Finally, Profile Generator outputs entity profiles for human reading. In the following sections, we say ``positive" entities if they match a label, or ``negative" entities if they do not.

\section{Generating Candidate Labels}
\label{sec:generate}
In our approach, we adopt a straightforward way to enumerate all candidate labels. The enumeration process creates labels by combining properties with all possible values. Indistinctive and trivial labels will be identified and filtered out by simple heuristic rules. 

\subsection{Building Candidate Pool}

Without any prior knowledge, these labels are enumerated brute-force from KG through an automatic process of label generation. It is straightforward to generate candidate \textit{AVL} and \textit{REL} labels by enumerating all combinations of attribute and values, or relations and entities. This process can be accomplished by one traversal through the entire KG.  Generating candidate \textit{AIL} and \textit{RAL} labels are more complex. Given a triple like $\left\langle ForrestGump, rating, 8.3 \right\rangle$, it is almost meaningless to simply generate a candidate label $\left\langle Film, rating, 8.3 \right\rangle$, because this label is too specific, not generalized, and can hardly represent typical features of other films. In our approach, we further generalize the continuous value of an attribute into a broader interval containing the value. The label $\left\langle Film, rating, [8.0,9.0]\right\rangle$ is better than $\left\langle Film, rating, 8.3 \right\rangle$, because the former is more representative for highly-rated films. 

Finding a proper interval for a label is basically a problem of discretization of continuous numerical value of the label. We set some simple discretization rules for finding proper intervals of some specific values. For example, we use a period of five years as the interval for all kinds of yearly values. For other kind of quantity values, the equal-width and equal-frequency are both simple and commonly-used discretization methods \cite{7}. However, the shortage of these methods is that they do not consider the data distribution. We adopt a local density-based discretization algorithm, which is rooted from \cite{8}. The major idea is to find the natural density interval of the attribute value. We ensure that the density in the middle part of the interval is high and the density near the boundary is low. After the attributive value is sorted, the density value shows a multi-peak phenomenon. Each peak of the density distribution indicates a boundary between two intervals. 

\subsection{Candidate Filtering}
The candidate pool may contain massive unqualified candidates representing trivial features. These labels provide very limited or even misleading information to the comprehension of entities. We consider two types of trivial candidates. The first type comprises \textbf{unrepresentative} labels. For example, in Drugbank\footnote{DrugBank: https://www.drugbank.ca/}, A candidate label $\left\langle Drug, accessID, ``DB00316"\right\rangle$ can be enumerated from the knowledge graph. But this label can only profile a single entity (a drug called Acetaminophen), but is  unrepresentative for other entities. Another source of unrepresentative labels comes from noisy data. In DBpedia, the \textit{birthDate} of some SoccerPlayer was stated in the year of 2915. These incorrect features lead to obviously meaningless labels. The other type of trivial candidates is \textbf{indistinctive} labels. If most of entities in a KG share a common feature, for example, the gender of all the Persons in a KG is female, users cannot distinguish entities from the label $\left\langle Person, gender, ``female"\right\rangle$. This type of labels provides near-zero information in entity comprehension. 

We use a simple heuristic rule to filter out trivial labels. Given an entity type $t$ and a candidate label $l$ related to $t$, we define $\mathcal{E}_t$ as the set of all entities of type $t$, and $\mathcal{E}^l_t \subseteq \mathcal{E}_t$ as the set of positive entities to the label $l$. A $support(l)$ is defined in ``\eqref{eq1},"  which is the ratio of positives in a population. $\alpha$ and $1-\alpha$ are the lower and upper threshold for non-trivial labels. We assume that labels with low $support$ are usually unrepresentative, while those with excessively high $support$ are indistinctive. Thus, a label in the middle of the spectrum is judged as a qualified candidate.

\begin{equation}
\alpha<support(l)=\frac{\left|\mathcal{E}^l_t\right|}{\left|\mathcal{E}_t\right|}<1-\alpha
\label{eq1}
\end{equation}

\section{Measuring Candidate Labels}
\label{sec:measure}
After candidate generation, all labels will go through a deeper investigation. The most important requirement is that they must be distinctive, characterizing a meaningful boundary between positives and negatives. In this section, we fully discuss the investigation on the distinctiveness of each candidate label.

\subsection{Distinctiveness of Labels}
While approaches of entity summarization usually look for informative features of entities, the most important task for entity profiling is to identify distinctive features. A good label should be able to distinguish a group of similar entities from other different ones. For example, a label representing highly-rated movies indicates that films of this type may have won the box-office or may be award-wining films. Films not-belonging to this type may perform differently. On the contrary, indistinctive labels are trivial, with which positive and negative entities will not show a remarkable difference. 

For a distinctive label, the positive entities are supposed to be similar, and meanwhile the negative entities are supposed to be diverse.  We use ``\eqref{eq2}," and ``\eqref{eq3}" to measure the distinctiveness. For a given label $l$ of type $t$, $\mathcal{E}^l_t$ denotes the set of positives of $l$, and $\mathcal{E}_t^{\bar{l}}$ denotes negatives. Entities in $\mathcal{E}^l_t$ should be similar with each other and they should also be dissimilar to those in the counterpart $\mathcal{E}_t^{\bar{l}}$. We define $d(l)$ as the degree of distinctiveness of $l$, which is the difference between the average internal similarity in $\mathcal{E}^l_t$ and the average external similarity from $\mathcal{E}^l_t$ to $\mathcal{E}^{\bar{l}}_t$. $sim(i,j)$ is the similarity between entity $i$ and entity $j$.

\begin{equation}
d(l)\!=\!avg\left(\!sim_{i,j\in \mathcal{E}^t_\tau}(i,j)\!\right)\!-\!avg\left(\!sim_{i\in \mathcal{E}^t_\tau,j\in \mathcal{E}_\tau^{\bar{t}}}(i,j)\!\right)\!
\label{eq2}
\end{equation}

\begin{equation}
d(l)=\frac{\sum_{i,j\in \mathcal{E}^l_t}sim_{i,j}(i,j)}{\left|\mathcal{E}^l_t \right|^2}-\frac{\sum_{i\in \mathcal{E}^l_t,j\in \mathcal{E}_t^{\bar{l}}}sim_{i,j}(i,j)}{\left|\mathcal{E}^l_t \right| \left|\mathcal{E}_t^{\bar{l}} \right|}
\label{eq3}
\end{equation}
\newline
\subsection{Measuring Distinctiveness by Entity Embedding}
Many methods have been proposed to measure entity similarities in a graph, such as Katz similarity \cite{9}, SimRank \cite{10}, and P-Rank \cite{11}. Their major ideas are: two entities are structurally similar if their neighbors are similar. So calculating similarity between two entities is transformed into a problem of iterative propagation of similarities along neighborhoods. However, there are two inevitable problems in these methods: 1) for large-scale KGs, path-based similarity measurements are not computationally feasible; 2) path-based methods are under homophily assumption, which is the tendency that entities interlinked with similar entities. However, more structural patterns can be shared by similar entities. As stated in \cite{12}, while two entities are not directly connected in KG, they are still similar if they are alike in many attributes, which is called attributive equivalency; or they act as similar structural role in the graph, which is called structural equivalency. Path-based methods are not able to measure these similarities.

We propose a multi-pattern entity embedding model to measure entity similarities, which is called the HAS model (HAS is an abbreviation of three path-finding strategies used in the model). A distributed representation of each entity is learned by HAS, preserving that entities are closed in a continuous low-dimensional space if they share one or more structural patterns. Three patterns are considered in HAS: homophily, attributive equivalency and structural equivalency. The HAS model simplifies the manipulation of entity representations, and it is effective and efficient in assessing distinctiveness for large-scale KGs. 

The idea of HAS model is inspired by the path-based embedding model, such as DeepWalk\cite{13} and Node2Vec\cite{14}, which use skip-gram \cite{15} to train the walk sequences of nodes in graphs and generate the vector of nodes. Distributed representation of nodes can be learned by modeling a stream of short random walks. These random walks encode features of the nodes and capture neighborhood similarity and community membership. However, DeepWalk only considers homophily pattern. Although Node2Vec considers the pattern of structural equivalency, it is still unclear which kind of similarity is encoded in any given path.

\begin{figure}[h]
\centering
\includegraphics[scale=0.3]{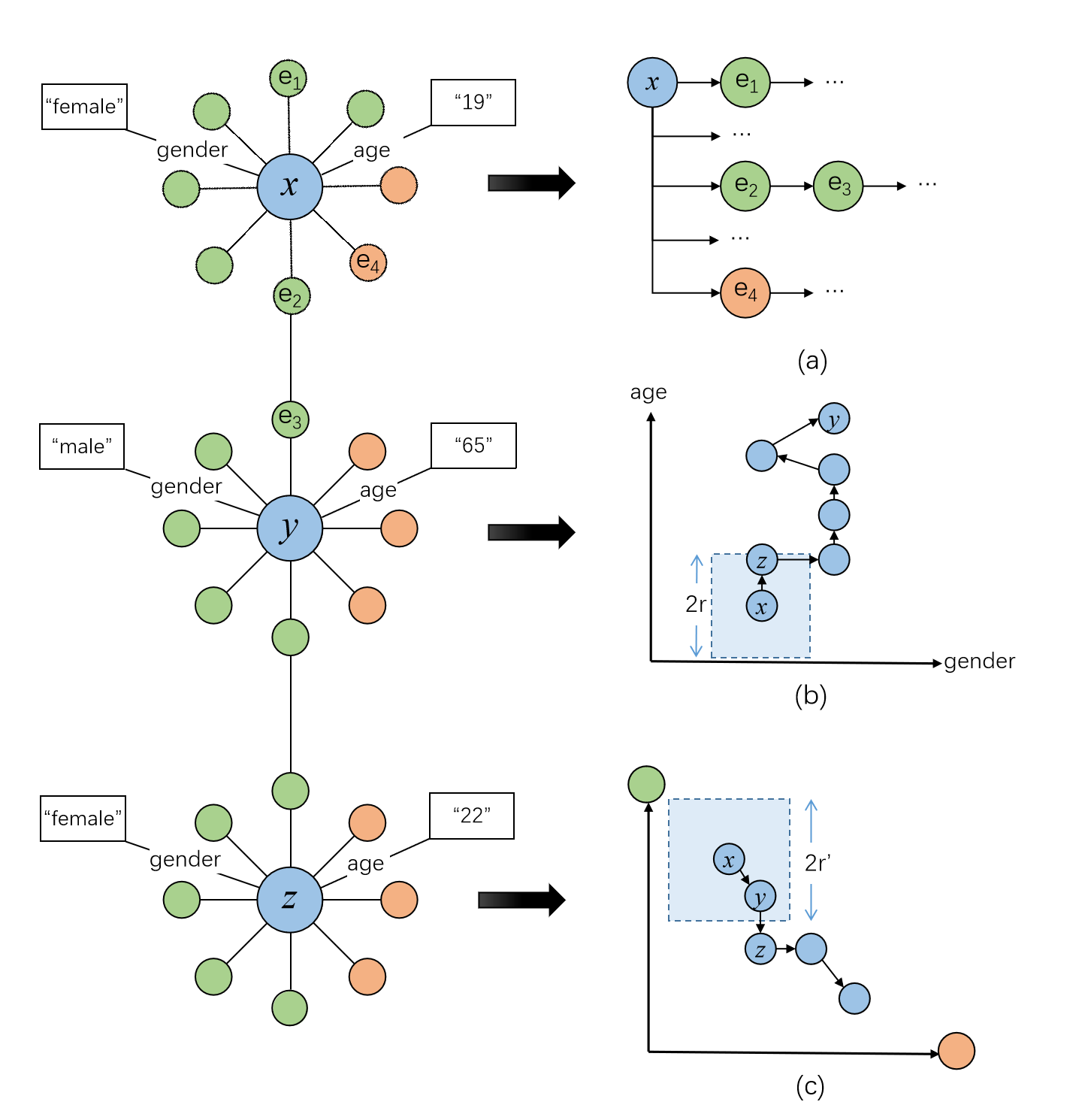}
\caption{Path finding for (a) H-path (b) A-path (c) S-path.}
\label{fig3}
\end{figure}

In our approach, given a KG as the input, a path finding operation is carried for each entity to find a set of paths starting from the entity. Three types of paths will be discovered by these strategies: (1) an H-strategy to discover H-paths representing homophily patterns; (2) an A-strategy to discover A-paths representing attributive equivalency; (3) an S-strategy to discover S-paths representing structural equivalency. Each type of path reflects a certain facet of the structural patterns. After Path Finding, these paths are proportionally mixed as entity features. Finally, we use skip-gram model to learn a continuous feature representation for each entity by optimizing a structural-pattern-preserving likelihood objective. Fig.\ref{fig3} shows three strategies of path finding. The left part of the figure illustrates a fragment of KG as input, where nodes are entities; white rectangles are literals; edges are relations or attributes linking entities and literals (we omit the directions of linking).

\vspace{0.5mm}

Entities are color-coded with their types. The right part of the figure shows three strategies of path finding starting from entity \textit{x}. We elaborate each strategy as following:

\textbf{H-Strategy}: H-Strategy is a simple random-walk strategy to find H-paths reflecting homophily patterns, which refers to direct connections between entities. In Fig.3(a), starting from \textit{x}, multiple paths can be generated simply using Deep-First-Search(DFS). 

\textbf{A-Strategy}: A-Strategy is used to find A-paths reflecting attributive equivalency. Starting from \textit{x}, A-Strategy tries to find a subsequent entity with the same type in the KG that is most similar to \textit{x} in attribute values. In Fig.\ref{fig3}, entities \textit{y} and \textit{z} are the same type with \textit{x}. We can tell that \textit{z} is more similar to \textit{x} than \textit{y}, because \textit{z}'s gender is the same with \textit{x} and \textit{z}'s age is close to \textit{x} comparing to \textit{y}. So in terms of attributive equivalency, A-Strategy will select \textit{z} as the subsequent node of a path starting from \textit{x}. Following this intuitive strategy, we can generate a set of paths starting from \textit{x}. However, in a large-scale KG, just to find a single closest neighbor as the next hop is not computational feasible, not to mention a set of multi-hop paths for all entities. 

Fig.3(b) shows a random walk model for finding A-paths. We first embed entities of the same type into an $\left| \mathcal{A}_t \right|$-dimensional attributive space, where $\left| \mathcal{A}_t \right|$ is the number of attributes of $t$. In this example, \textit{x}, \textit{y} and \textit{z} are embedded into a 2D space (age and gender). After a normalization on each dimension, a virtual square (a hypercube in case of multi-dimensional space) taking \textit{x} as center outlines the close neighbors of \textit{x}. We denote the edge length of the square as $2r$. Entities falling in the area of the square are seen as sufficiently close neighbors of \textit{x}. A random selection will select one of the close neighbors as the consequent entity of \textit{x} in an A-path. Iteration continues until a fix-length A-path is generated for \textit{x}. The initial setting of $2r$ is estimated based on the average interval between adjacent entities in attributive space. The hypercube can be zoomed in or zoomed out so that the number of close neighbors in the hypercube approaches the average number of direct neighbors (average degree) in the original KG. 

\textbf{S-Strategy}: Structural equivalency is usually embedded in the local structures of entities. For example, two professors have a high structural equivalency if they play similar roles in their social networks, for example each of them connects to many students. Similar to the A-Strategy, S-Strategy finds S-path by first embedding entities in a structural space. Given a type $t$, its structural space has $\left| \mathcal{T} \right|$ dimensions, where $\left| \mathcal{T} \right|$ is the number of types in a KG. The component of an entity in a certain dimension $t’$ is the number of its direct neighbors of type $t’$. In Fig.3(c), the horizontal axis represents the number of neighbors typed as orange, and the vertical axis represents green. The coordinates of \textit{x}, \textit{y} and \textit{z} are $\left\langle 2,6\right\rangle,\left\langle3,5 \right\rangle,\left\langle3,4 \right\rangle$ respectively. The following steps of path finding are quite similar to the ones in A-Strategy. 

\textbf{Path Mixing}: Finally, for each entity, three set of random-walk-based paths $P^H, P^A, P^S$ are generated by HAS model, where $P^H, P^A, P^S$ stand for H-path, A-path and S-path respectively. These paths will be sampled into a final feature set. Shown in ``\eqref{eq4}," $P$ is the final feature set of an entity, $\lambda_H,\lambda_A,\lambda_S$ are proportional parameters for path sampling. A uniformed path sampling is a strategy with $\lambda_H,\lambda_A,\lambda_S=1:1:1$. A bias sampling is a strategy with an unbalanced weighting scheme. Especially, when $\lambda_H,\lambda_A,\lambda_S=1:0:0$, HAS model is identical to DeepWalk. After the path mixing, we follow the Skip-gram learning process of DeepWalk, and here we omit the details. 

\begin{equation}
P=\lambda_HP^H\cup \lambda_AP^A\cup \lambda_sP^S
\label{eq4}
\end{equation}

\subsection{Label Set and Entity Profiles}
Before adding distinctive labels into the final label set, we need to do the last investigation. A good label should satisfy two requirements: (1) it brings little redundancy to the label set; and (2) it increases the completeness of the label set. The first requirement prioritizes labels that are different with already-selected labels. The second requirement favors labels that are complementary to the already-selected labels.

We propose a re-ranking method to finally investigate candidate labels. As shown in ``\eqref{eq5}," given a candidate label $l_i$ that is in a candidate label set $\mathbb{L}^c_t$ but not in the final label set $\mathbb{L}_t$ yet, $d(l_i)$ is the distinctiveness score of $l_i$, $reward(l_i, \mathbb{L}_t)$ is the potential contribution of $l_i$ to the increase of the total coverage of positive entities in the KG. $penalty$ is the potential impact of $l_i$ to the increase of redundancy in $\mathbb{L}_t$. The definition of the reward and penalty is defined in ``\eqref{eq6}," and ``\eqref{eq7}." $\delta$ is a bias factor. $\mathcal{E}_t$ is the set of entities of type $t$, and $\mathcal{E}_t^{l_i}$ stands for the positive entities of the label $l_i$. Finally, candidate labels are ranked and picked up one by one into the label set.

\begin{equation}
\begin{split}
l_i=&\mathop{\arg \max}_{l_i\in {\mathbb{L}^c_t }}[d(l_i)+\delta \cdot reward(l_i,\mathbb{L}_t)\\
&-(1-\delta)penalty(l_i,\mathbb{L}_t)]
\end{split}
\label{eq5}
\end{equation}

\begin{equation}
reward(l_i,\mathbb{L}_t)=\frac{\left| \bigcup _{l_j\in (\mathbb{L}_t\cup \{l_i\} )}\mathcal{E}_t^{l_j}\right|}{\left|\mathcal{E}_t\right|}
\label{eq6}
\end{equation}

\begin{equation}
penalty(l_i,\mathbb{L}_t)=\frac{\sum_{l_j\in \mathbb{L}_t}\left|\mathcal{E}_t^{l_i}\cap \mathcal{E}_t^{l_j}\right|}{\left|\mathbb{L}_t\right|\times \left|\mathcal{E}_t\right|}
\label{eq7}
\end{equation}

It is straightforward to generate entity profiles with the label set. All the entity descriptions will be scanned to see whether an entity matches certain labels. At last, entity profiles will be presented to users to promote a rapid comprehension on the uniqueness of entities. 

\section{Evaluation}
\label{sec:evaluation}
In this section, we introduce the datasets in our experiment. Two experiments are described: an intrinsic evaluation on the quality of the label set and the entity profiles, and an extrinsic evaluation to verify the usefulness of our approach in a practical task. 

\subsection{Datasets and Settings}
We have published the source code of our project on github\footnote{https://github.com/wds-seu/enprofiler}. In our experiments, we use two real-world knowledge graph to evaluate entity profiling: (1) DBpedia\footnote{https://wiki.dbpedia.org/downloads-2016-10},which is a domain-independent encyclopedic dataset covering a broad range of descriptions of entities, such as people, location and company. (2) LinkedMDB\footnote{http://www.linkedmdb.org/},which is a knowledge graph about films, actors, director, and other entities in film industry. Some statistics of the datasets are shown in Table \ref{tab2}. 
\begin{table}[h]
\centering
\caption{Statistics of DBpedia and LinkedMDB}
\vspace{2mm}
\setlength{\tabcolsep}{4pt}
\begin{tabular}{|p{60pt}|p{70pt}<{\centering}|p{70pt}|}
\hline
            & \textbf{DBpedia} & \textbf{LinkedMDB} \\ \hline
\#triple    & 38,285,143       & 6,148,119          \\ 
\#type      & 422              & 53                 \\ 
\#entity    & 5,150,432        & 740,469            \\ 
\#relation  & 18,746,174       & 1,211,046          \\ 
\#attribute & 14,388,537       & 3,274,855          \\ \hline
\end{tabular}
\label{tab2}
\end{table}

In our implementation, each entity is represented by a 200-dimensional vector using HAS model. The feature set of each entity includes 100 random-walk-paths of maximum 8 hops. Other hyper-parameters are empirically set as follows: $\alpha$ is set to 0.1, and the bias factor $\delta$ is set to 0.5.

\subsection{Evaluation on the Label Set}
To evaluate the label quality, we select a subset in each dataset to generate a ground-truth label set. In DBpedia, we select 15 types of entities including ``Airline", ``Band", ``BaseballPalyer", ``Lake", ``University", ``Philosopher", ``Song", ``PoliticalParty", ``TelevisionShow", ``Comedian", ``AcademicJournal",  ``Actor",  ``Book",  ``Mountain", and ``RadioStation", and we also select ``Film" In LinkedMDB. The main reason we pick out these types is for the convenience of human experts to efficiently generate a ground truth. Selected types of entities have relatively abundant information, and they are easy to understand by human experts. Five experts on knowledge graph were invited to read through the subsets and to manually construct a label set containing top-5 and top-10 distinctive labels for each type of entities. To reduce the human effort, labels in the ground-truth label set are simplified labels that only contain the property part of the label. Experts show a consensus on judging distinctive labels. The average agreements between ground-truth label sets are 2.87 and 6.09 for top-5 and top-10 label sets respectively. 

For the selected types, the size of the model-generated label sets is shown in Table \ref{tab3}, where (c), and (f) stand for the number of enumerated candidate labels and the selected labels after filtering. It is obvious that our model abandoned a large number of trivial candidates. In DBpedia, attributive information of entities is not in abundance comparing to relational information, which leads to a relatively small number of $AIL$, and $AVL$ labels in the label set. In LinkedMDB, there is more attributive information for ``Film", but all the attributive information are trivial and unrepresentative, such as the id and link of the films. They are eliminated by the filtering process. Thus, the $AIL$, $AVL$, and $RAL$ of LinkedMDB become empty. 

\begin{table}[h]
\caption{Statistics of the Model-generated Label Set}
\vspace{2mm}
\setlength{\tabcolsep}{2pt}
\begin{tabular}{|p{80pt}|p{35pt}<{\centering}|p{35pt}<{\centering}|p{35pt}<{\centering}|p{35pt}<{\centering}|}
\hline
\textbf{Dataset} & \textbf{\#\textit{AIL}} & \textbf{\#\textit{AVL}} & \textbf{\#\textit{RAL}} & \textbf{\#\textit{REL}} \\ \hline
\textbf{DBpedia(c)} & 625           & 419,940             & 389,906          & 904,423          \\ 
\textbf{DBpedia(f)} & 143            & 54             & 1,521          & 2,203          \\ \hline
\textbf{LinkedMDB(c)}   &   10          &  181,648               & 5              &  13,345            \\ 
\textbf{LinkedMDB(f)}   & /              & /              & /              & 176      \\ \hline
\end{tabular}
\label{tab3}
\end{table}
While only a subset of our datasets is used for the manual generation of the ground truth, the complete DBpedia and LinkedMDB are used for model training. In generating a label set, we compete our methods with several baselines defined as following:
\newpage
\begin{itemize}
\item \textbf{Random}: one of the two baselines, which generates a label set by a random selection on all candidate labels;

\item \textbf{TF-IDF}: the other baseline, which generates a label set by using TF-IDF to measure the importance of a label. This method is often adopted as a baseline in entity summarization;

\item \textbf{Filtering}: our method using support value to filter out trivial labels; 

\item \textbf{H/A/S}: our method using a separate H-Strategy / A-Strategy / S-Strategy for entity embedding;

\item \textbf{HAS}: our method using a complete HAS model for entity embedding;
\end{itemize}

In Table \ref{tab4} and Table \ref{tab5}, we evaluate the performance of our model against baselines by calculating the MAP value and F-measure. On each metric, we compare the agreement between top-5/10 model-generated labels and the ground truths. All our methods outperform the baseline Random. Our method of Filtering is better than TF-IDF on DBpedia, but not as good on LinkedMDB. It shows that just by heuristic filtering on candidate labels is not enough to produce a high-quality label set. On DBpedia, H-Strategy has a better performance comparing to A- and S-Strategy on identifying distinctive labels. This is because there is plenty of relational information for the selected types of entities in DBpedia. The performance of H/A/S varies on LinkedMDB. A-strategy is not applicable on LinkedMDB because all attributive information is eliminated by filtering process due to its triviality. H-strategy significantly outperforms S-strategy. The results clearly show that a biased weighting scheme is needed for HAS model, which will be studied in our future work. 

\begin{table}[h]
\caption{MAP and F-measure of the Label Set of DBpedia}
\vspace{2mm}
\setlength{\tabcolsep}{2pt}
\begin{tabular}{|p{40pt}|p{42pt}<{\centering}|p{40pt}<{\centering}|p{50pt}<{\centering}|p{46pt}<{\centering}|}
\hline
\textbf{Method}&	\textbf{MAP@5}&	\textbf{F-M@5}&	\textbf{MAP@10}&	\textbf{F-M@10}\\
\hline
Random	&0.105	&0.184	&0.142	&0.281\\
TF-IDF	&0.109	&0.197	&0.167	&0.296\\
\hline
Filtering	&0.118	&0.243	&0.245	&0.410\\
H	&\textbf{0.180}	&0.296	&0.278	&0.426\\
A	&0.115	&0.208	&0.228	&0.363\\
S	&0.100	&0.178	&0.235	&0.401\\
\hline
HAS	&0.175	&\textbf{0.298}	&\textbf{0.295}	&\textbf{0.424}\\
\hline
\end{tabular}
\label{tab4}
\end{table}

\begin{table}[h]
\caption{MAP and F-measure of the Label Set of LinkedMDB}
\vspace{2mm}
\setlength{\tabcolsep}{2pt}
\begin{tabular}{|p{40pt}|p{42pt}<{\centering}|p{40pt}<{\centering}|p{50pt}<{\centering}|p{46pt}<{\centering}|}
\hline
\textbf{Method}&	\textbf{MAP@5}&	\textbf{F-M@5}&	\textbf{MAP@10}&	\textbf{F-M@10}\\
\hline
Random	&0.036	&0.133	&0.034	&0.200\\
TF-IDF	&0.067	&0.190	&0.071	&0.260\\
\hline
Filtering	&0.045	&0.183	&0.052	&0.209\\
H	&0.253	&0.440	&\textbf{0.283}	&\textbf{0.440}\\
A	&-	&-	&-	&-\\
S	&0.210	&0.320	&0.239	&0.400\\
\hline
HS	&\textbf{0.280}	&\textbf{0.450}	&0.263	&0.380\\
\hline
\end{tabular}
\label{tab5}
\end{table}

In DBpedia, the complete HAS model has the best F-measure@10, followed by a standalone H-Strategy, while the best F-measure@10 on LinkedMDB is generated by H-Strategy. It implies that the inner structure and the information richness of the dataset have a great impact on entity profiling. For some datasets, some strategies may have less impact on the entity profiling, such as A- and S-strategy. We also evaluate the impact of the re-ranking on the quality of labels. In Table \ref{tab6}, the F-measure@10 performance for each selected type is listed, where HAS+r represents our method with both HAS model and re-ranking. It is clear that for most of entity types, the performance of our model improves remarkably with the help of re-ranking, which shows that the model-generated label set matches more with human judges by redundancy reduction and coverage promotion. Averagely, the re-ranking paradigm improves the quality of the label set by 30$\%$.

\begin{table}[h]
\caption{F-measure@10 Performance with Re-ranking}
\vspace{2mm}
\setlength{\tabcolsep}{2pt}
\centering
\begin{tabular}{|p{50pt}|p{50pt}<{\centering}|p{50pt}<{\centering}|p{50pt}<{\centering}|}
\hline
  & \textbf{Types}      & \textbf{HAS} & \textbf{HAS+r} \\ \hline
\textbf{L.MDB}                     & Film  & 0.38         & \textbf{0.44}           \\ \hline
\multirow{15}{*}{\textbf{DBpedia}} & Airl. & 0.402        & \textbf{0.424}          \\ 
                                   & Band  & 0.26         & \textbf{0.56}           \\  
                                   & Base. & 0.46         & \textbf{0.7}            \\  
                                   & Lake  & 0.28         & \textbf{0.4}            \\ 
                                   & Univ. & 0.177        & \textbf{0.406}          \\  
                                   & Phil. & 0.288        & \textbf{0.667}          \\ 
                                   & Song  & 0.538        & \textbf{0.807}          \\ 
                                   & Poli. & 0.209        & \textbf{0.524}          \\ 
                                   & TVsh. & 0.186        & \textbf{0.478}          \\ 
                                   & Come. & 0.528        & \textbf{0.575}          \\ 
                                   & Acad  & \textbf{0.84}         & \textbf{0.84}           \\ 
                                   & Acto. & 0.36         & \textbf{0.42}           \\  
                                   & Book. & \textbf{0.6}          & \textbf{0.6}            \\ 
                                   & Moun. & 0.609        & \textbf{0.645}          \\ 
                                   & Radi. & \textbf{0.62}         & 0.532          \\ \hline
\multicolumn{2}{|c|}{\textbf{Average}}     & 0.421        & \textbf{0.566}          \\ \hline
\end{tabular}
\label{tab6}
\end{table}
 
\subsection{Evaluation on Entity Profiling} 
It is hard for human experts to produce ground-truth profiles for the entities under evaluation. Thus, in the evaluation of profile qualities, we change the evaluation mechanism. Ten experts on Knowledge Graph were invited to rate the model-generated entity profiles. We selected 10 entities for each type and generated their profiles using our model. For each entity, its profile comprises 1 to 5 labels. Every expert was presented with the model-generated profiles, together with the original triple description of each entity and some statistical indications, like the support value of the labels, the number of entities in the type, etc. Each profile would be manually rated between 0 to 2 points: 0 for indistinctive profiles that provide little information for entity comprehension; 1 for borderline profiles; 2 for distinctive profiles that are helpful for entity comprehension. To alleviate the burden of human ranking, we chose 5 types of entities to evaluate the profile quality. The average ratings are shown in Table \ref{tab7}, in which HAS model with re-ranking has the best performance among all methods.

\begin{table}[h]
\caption{Human Rating on the Quality of Entity Profiles}
\vspace{2mm}
\setlength{\tabcolsep}{1pt}
\centering
\begin{tabular}{|p{40pt}|p{25pt}<{\centering}|p{25pt}<{\centering}|p{25pt}<{\centering}|p{28pt}<{\centering}|p{25pt}<{\centering}|p{22pt}<{\centering}|p{25pt}<{\centering}|}
\hline
                & \multicolumn{5}{c|}{\textbf{DBpedia}}                                    &\multicolumn{2}{c|}{\textbf{L.MDB}}                \\ \hline
\textbf{Method} & Acad.        & Acto.        & Book         & Moun.        & Radi.        & Film           & \textbf{Avg.} \\ \hline
Random          & 0.2          & 0.6          & 0.6          & 1            & 0.8          & 0.6            & 0.6          \\ 
TF-IDF          & 0.8          & 1.1          & 1.3          & 0.8          & 0.9          & 1              & 1             \\ \hline        
Filtering       & 0.5          & 0.8          & 0.9          & 0.7          & 0.8          & 0.9            & 0.8  
  \\
H               & 1.4          & \textbf{1.4} & \textbf{1.4} & 1.3          & \textbf{1.6} & 1.4            & 1.4           \\ 
A               & 1.5          & 0.8          & \textbf{1.4} & \textbf{1.5} & 1.5          & /              & 1.4           \\ 
S               & 1.4          & 1.2          & 1.3          & 1.4          & 1.5          & 1              & 1.3           \\ \hline
HAS+r           & \textbf{1.6} & 1.3          & \textbf{1.4} & \textbf{1.5} & \textbf{1.6} & \textbf{1.5}   & \textbf{1.5}  \\ \hline
\end{tabular}
\label{tab7}
\end{table}

\subsection{Extrinsic Evaluation on Entity Profiling}
We design a \textit{spot-the-difference} game for the extrinsic evaluation on profile quality. In this game, we manually construct 20 multiple-choice questions, in which each question contains four entities as options. The trick is: in each question there is one and only one entity that is quite different from the others. For example, a question contains four pop-music Asian singers from DBpedia: \textit{Coco Lee}, \textit{Mc Jin}, \textit{Stanley Huang}, and \textit{Hebe Tien}. The different one is \textit{Hebe Tien} for that: from DBpedia, although they share much commons like the race and the professions, the music genre of \textit{Hebe Tien} is Jazz, while the genres of others are all hip-hop. Ten players were invited to play this game, and they were required to spot the different entity accurately and quickly. We divided the players into two groups: an experimental group (EG), in which each player was provided with the profiles of the entities, together with the original triple descriptions; a controlled group (CG), in which only original triple descriptions were provided. We examined: (1) the average accuracy of the players; and (2) the average time they spent in each question. The result of the game is shown in Table \ref{tab8}. We observe that the experimental group shows a higher accuracy (16\% improvement) in the game and consumed significantly shorter time (2.2 times faster). Compared with the lengthy and complicated triple descriptions, entity profiles are informative and distinguishing, which remarkably helped the players to spot the difference quickly.

\begin{table}[h]
\caption{Accuracy and Time Performance of Players in Different Groups}
\vspace{2mm}
\setlength{\tabcolsep}{2pt}
\centering
\begin{tabular}{|p{65pt}|p{25pt}<{\centering}|p{25pt}<{\centering}|p{25pt}<{\centering}|p{25pt}<{\centering}|p{25pt}<{\centering}|p{25pt}<{\centering}|}
\hline
                    & \multicolumn{5}{c|}{\textbf{Results}} & \textbf{Avg.}  \\ \hline
\textbf{Acc(EG,\%)} & 95    & 95    & 85    & 80    & 85    & \textbf{88}    \\ \hline
\textbf{Acc(CG,\%)} & 70    & 80    & 75    & 70    & 75    &  72    \\ \hline
\textbf{Time(EG,m)} & 8.05  & 7.03  & 10.35 & 11    & 9.15  & \textbf{9.12}  \\ \hline
\textbf{Time(CG,m)} & 27.51 & 17.49 & 18.29 & 18.21 & 18.29 &  19.96 \\ \hline
\end{tabular}
\label{tab8}
\end{table}
\subsection{Discussion on the Knowledge Incompleteness}
\begin{figure}[h]
\centering
\includegraphics[scale=0.3]{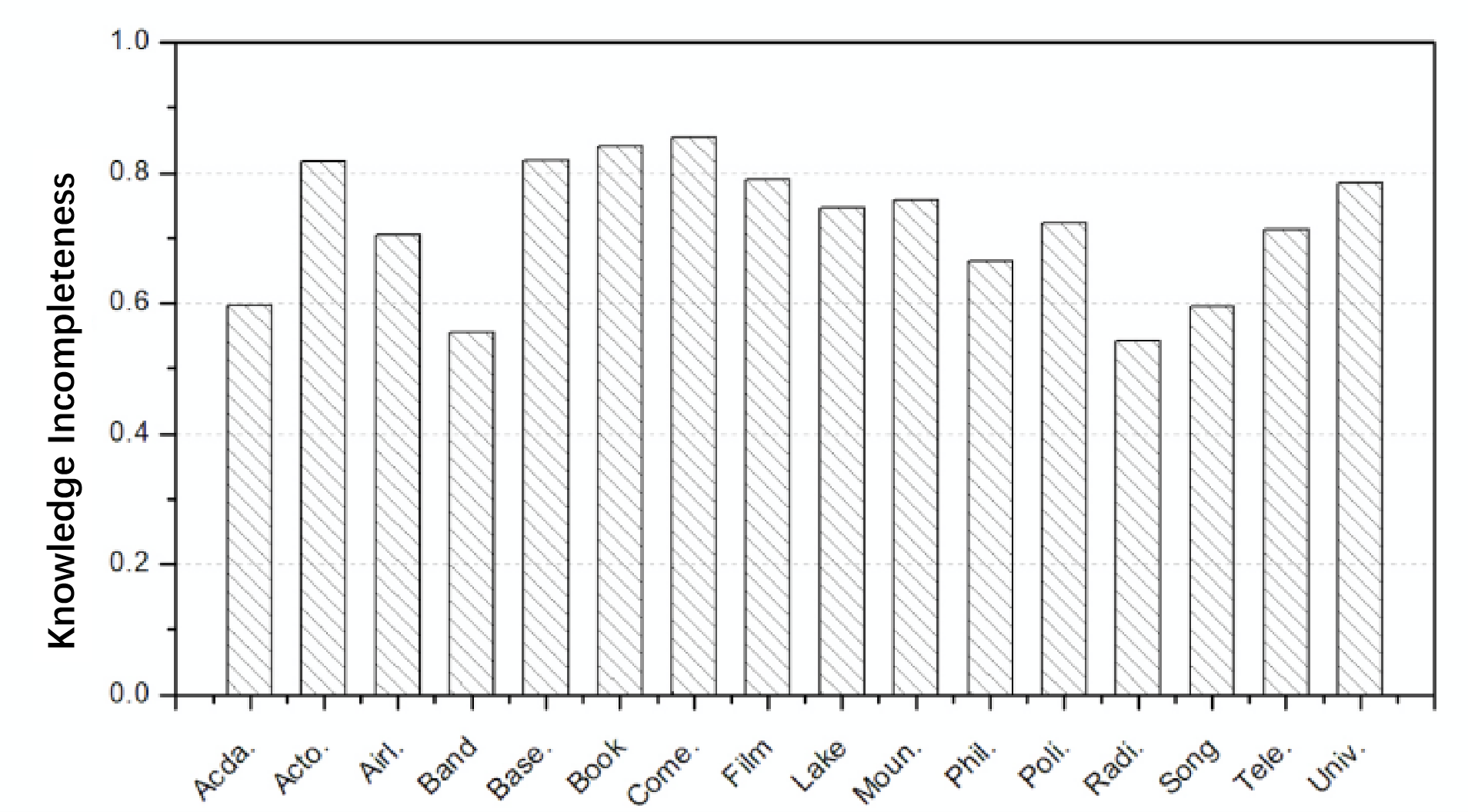}
\caption{Average knowledge incompleteness of selected types of entities.}
\label{fig4}
\end{figure}

Knowledge incompleteness is a common issue in many large-scale KGs. For example, some film entities in LinkedMDB have specified filming locations, while many other film entities lack a definition on this attribute. In Fig.\ref{fig4}, We illustrate the degree of knowledge incompleteness for each selected type. This problem has a negative impact on entity profiling. For those incompletely-defined entities, our system usually produces sparse profiles, which consist of few labels and are not informative for users. It is our future work to deal with knowledge incompleteness in entity profiling. 

\vspace{0.5mm}

\subsection{Comparison to Entity Summarization}

To analyze the pros and cons between the two paradigms of Entity Summarization and Entity Profiling, we illustrate the intuitive difference between these paradigms. We conducted an experiment comparing our approach against FACES \cite{21}, which is a successful and representative work about entity summarization. FACES provides model-generated summaries for its benchmark dataset, in which each entity is summarized with top-5 and top-10 descriptions. We invite 10 experts with background in Semantic Web to compare the helpfulness of both paradigms. 15 entities are selected from the benchmark dataset, including Actors, Cities, Companies, Countries and Films in DBpedia. They are manually summarized by the experts. 

In Table \ref{tab9}, the MAP value and F-measure performance for each paradigm are listed. Our method outperforms FACES in the top-5 experiment. It shows that the approach of entity profiling is more close to human understanding comparing to short entity summaries. However, our method falls behind on F-measure@10. This is because in the selected entities, some are described with few distinctive information, which leads a short profile with less than 10 labels. In these cases, the agreement between the profiling approach and the human ground truth decreases. This is also a reflection on the negative impact of knowledge incompleteness. \begin{table}[h]
\caption{MAP and F-measure of our Approach against FACES on 15 Entities}
\vspace{2mm}
\setlength{\tabcolsep}{2pt}
\centering
\begin{tabular}{|p{40pt}|p{42pt}<{\centering}|p{40pt}<{\centering}|p{45pt}<{\centering}|p{45pt}<{\centering}|}
\hline
\textbf{Method} & \textbf{MAP@5} & \textbf{F-M@5} & \textbf{MAP@10} & \textbf{F-M@10} \\ \hline
\textbf{FACES}  & 0.155          & 0.211          & 0.217           & \textbf{0.403}  \\ 
\textbf{HAS+r}    & \textbf{0.260} & \textbf{0.375} & \textbf{0.233}  & 0.295           \\ \hline
\end{tabular}
\label{tab9}
\end{table}

\begin{figure}[h]
\centering
\includegraphics[scale=1.0]{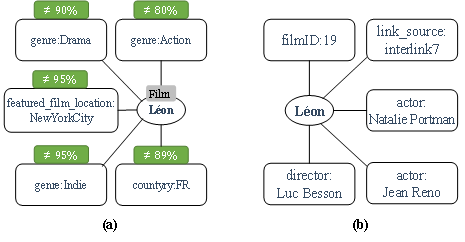}
\caption{An example of entity profiling vs. entity summarization.}
\label{fig5}
\end{figure}

Intuitively shown in Fig.\ref{fig5}, on the left side of the figure is the profile of the film entity ``$L\acute{e}on$", which is already shown as an example in Fig.1. On the right side is a ground-truth summarization (which is provided in the 1st International Workshop on Entity Retrieval \footnote{ https://sites.google.com/view/eyre18/home }) of the same entity with five extracted statements. While the summary of the film clearly describes its critical information, such as filmID and link sources, our entity profiling focuses more on the features that distinguish this film from others, such as its genre, country, and filming locations. We believe entity profiling is a good complement to entity summarization in various knowledge graph tasks. 

\vspace{0.5mm}

\section{Related Works}
\label{sec:related works}
The entity profiling has gradually gained the attention of the academic community in recent years, but in general it has not been fully studied. In \cite{16}, an entity profiling approach was introduced to characterize real world entities that may be described by different sources in various ways with overlapping information. The authors focused their goal on building a complete and accurate picture for a real world entity  despite possibly conflicting or even erroneous values. In \cite{17},  a corpus was introduced to profile entities in microblog. The corpus helps to profile the entities by annotating topics and opinions in microblog. However, our target is interrelated entities in the KG rather than entities in natural language text. In \cite{18},  entity profiling was proposed for experts to identify the expertise of each expert. They considered that entity profiling has significant dynamic characteristics, which means the result of entity profiling would change over time. It was a good idea to mine dynamic time and space features in entity profiling. 

Besides, we can find two similar research fields. One category of related works is about a fundamental profiling task: user profiling, which is similar to ours in that many user profiling approaches also adopt a paradigm to abstraction labels from user data. The other category is entity summarization, which usually adopt an extractive paradigm. But the motivation of entities summarization is highly related to ours, that is to promote entity comprehension. User profiling was divided into two types in \cite{19}: mobility profiling and demographic profiling from multi-source data, such as images, texts, behaviours from social media. Mobility profiling referred to dynamic space-time characteristics. Demographic profiling \cite{20} included some static characterization for entities, such as age, gender. The data source was very rich for profiling in their work. The other profiling work was based on ontology \cite{21}, using association rules algorithm to mine rich profiling information from user attributes behaviour. For the task of entity summarization, most of researches are categorized as diversity-centred summarises \cite{22,23} and relevance-oriented summaries\cite{3,24,25}. The former took the comprehensiveness and diversity of summaries into consideration. For the latter one, the importance of the connected resource and the relevance for the target entity are prioritized.

\section{Conclusions and Future Works}
\label{sec:conclusions}
In this paper, we propose entity profiling, which is a new paradigm to help user understand entities in KGs. Different with extractive entity summarization, entity profiling is an abstractive approach, which focuses on automatically discovering distinctive labels from KG to profile entities. The goal of profiling is to help users understand the uniqueness of entities among others. We propose a HAS model and a re-ranking method to deeply investigate the distinctiveness of labels in profiling. We conducted intrinsic and extrinsic experiments on real dataset to validate the effectiveness of entity profiling. To the best of our knowledge, we are the first to propose the research of entity profiling in knowledge graph, and we believe entity profiling is a good complement to the research of entity summarization to fulfill various knowledge graph tasks.

In our future works, we will fully investigate the issue of knowledge incompleteness. A joint model on knowledge completion and entity profiling will be proposed. To further validate the effectiveness of entity profiling, we will attempt to leverage entity profiling in downstream tasks, such as entity linking and entity recommendation, in which the comprehension of entities is critical for the fulfillment of these tasks.

\end{document}